"Medical robotics: where we come from, where we are and where we could go"
Jocelyne Troccaz, CNRS Research Director
TIMC laboratory – CNRS – Grenoble University and Hospital

Medical robotics can be split in subfields depending on who has to be assisted by a robot; it can be:
- The hospital: for instance for patient transportation and handling,
- the patient: by providing him/her robotized prostheses or artificial organs, rehabilitation aids or assistance for every day tasks,
- the clinician: by assisting his diagnostic or therapeutic tasks.

This short note focuses on the third type of systems which purpose is to provide the clinician with robotic aids for moving and actuating medical sensors and tools in minimally invasive interventions. Such robots were introduced in the early eighties first in neurosurgery applications, then, a little later, for orthopaedics; the aim was to transfer accurate machining and positioning capabilities from the industrial world to the clinical one. However one major difference with industrial robotics is the need for interaction with a clinical user. Different types of systems have been proposed depending on the level of autonomy they leave to the robot. In the first generation of robots, two main types of interactions were implemented: active robots execute in an autonomous way subtasks under human supervision ("the automation dream") whilst semi-active robots position a tool-guide which is used by the clinician to carry out his task, for instance introducing a needle into the body. Because the clinical environment is very complex and constrained, because the medical knowledge is very large and difficult to model exhaustively and finally because the clinician is an expert with high abilities to detect, analyze and react to unwanted critical situations, the automation dream has often turned into a collaborative framework where both the clinician and the robot have to work together in a synergistic way. Thus more recently, new interaction paradigms were proposed: co-manipulation consists in having a tool attached to the robot end-effector and held by the clinician; both of them contribute to its motion in a programmable way depending on the task execution status. Tele-operation approaches, involving master-slave architectures were also introduced for complex interventions on movable and deformable organs and for microsurgery applications. The automation paradigm has been recently re-explored in order to synchronize tool motions to organ ones in specific applications (radiotherapy in particular). Robot architectures also evolved this last decade from multi-purpose large robotic arms to more dedicated and small robots.

Medical robotics raises specific issues related to the application field. For instance, a robot that carries a surgical tool has to be at least partly cleanable as any instrument in contact with the patient for asepsis. This may have consequences on design choices. Since the robot moves in close proximity to people – the medical staff and the patient – and may manipulate invasive tools, robustness and safety are critical issues. Several hardware and software engineering solutions can contribute to solving them. Developing small specific robots whilst limiting the application scope makes the integration to the clinical environment often simpler: man-machine interaction is easier; modifications in the clinical workflow are more limited; workspace is more controllable; safety is more easily demonstrated, etc.

The medical robot has to face a very serious competitor: the navigation system. Surgical navigation is based on the real-time intra-operative localization of objects (tools, sensors, anatomical structures) for instance with an optical system tracking markers placed on the object to be localized. It provides to the clinician information about his/her actual task with respect to the planned one and/or relatively to non visible anatomical structures. Such navigation systems have been largely used in clinics (neurosurgery and orthopaedics in particular). Their integration in the clinical environment is generally simpler than the integration of robots; safety is easier to assess; cost is often lower than robots' one.

Although the field of medical robotics has produced lots of research results and many prototypes, rather few systems really entered the operating theatre and demonstrated their clinical usefulness. Indeed, one major difference with other applications of robotics is the need for the proof of *clinical added-value* of the medical robot. In that context, demonstrating that the robot can safely and accurately machine a bone surface or insert a screw is not sufficient: one must prove with quantitative data that the system has a clinical advantage over other existing techniques. The advantage may be in less complications, reduced hospitalization time and cost, higher diagnostic efficacy, etc. Such a demonstration requires careful evaluation involving series of patients in different centres and is regulated by laws and ethical codes which depend on the countries where it takes place. When the expected advantage is a long-term one, for instance the stability and behaviour of a knee or hip prosthesis, this evaluation can last more than ten years raising obvious economical issues. Developing medical robots thus appear to be a complex, expensive and long process which requires a constant collaboration of clinics, academic research and industrial development. The interaction between those different actors has to start in the very early phases of the project in order to find cost-effective, innovative technical solutions that are compatible with a clinical use and that may lead to significant breakthroughs.

The future of medical robotics is probably in the augmentation of surgical instruments with sensors or actuators rather than in placing a traditional tool on the robot end-effector. The time has come for large series of MEMS devices and disposable systems; smart pills for gastric track exploration are already in clinical use. With such technology the spectrum of applications is likely to expand in a dramatic way.